\documentclass[11pt]{article}
\usepackage{fullpage}
\usepackage{authblk}
\newif\ifdraft
\drafttrue
\draftfalse

\usepackage{xspace}
\usepackage[table,svgnames,dvipsnames]{xcolor}               %Must be before hyperref

\definecolor{Darkgreen}{rgb}{0,0.4,0}
\definecolor{Darkred}{rgb}{0.8,0,0}
\definecolor{Darkblue}{rgb}{0,0,0.4}
\definecolor{Gray}{gray}{0.9}
\definecolor{BekirBlue}{rgb}{0.18,0.33,0.59}

\newcommand{\bekir}[1]{}
\newcommand{\yan}[1]{}
\newcommand{\ahmet}[1]{}
\newcommand{\marek}[1]{}
\newcommand{\nirav}[1]{}
\newcommand{\sindy}[1]{}

\ifdraft
  \renewcommand{\bekir}[1]{{\color{Maroon} Bekir says: #1\xspace}}
  \renewcommand{\yan}[1]{{\color{Violet} Yan says: #1\xspace}}
  \renewcommand{\ahmet}[1]{{\color{Darkblue} Ahmet says: #1\xspace}}
  \renewcommand{\marek}[1]{{\color{Darkred} Marek says: #1\xspace}}
  \renewcommand{\nirav}[1]{{\color{Darkgreen} Nirav says: #1\xspace}}
\fi
\renewcommand{\sindy}[1]{{\color{Magenta} Sindy says: #1\xspace}}    

\usepackage[backend=biber,style=numeric,citestyle=numeric]{biblatex} 
\usepackage{amsmath,amssymb}
\usepackage{lmodern}

\usepackage{iftex}

\usepackage{verbatim}
\usepackage{fancyvrb}
\usepackage{fvextra}

\usepackage{floatflt}
\usepackage{wrapfig, framed, caption}
\usepackage{subcaption}

\usepackage{tabularx}

\ifPDFTeX
  \usepackage[T1]{fontenc}
  \usepackage[utf8]{inputenc}
  \usepackage{textcomp} % provide euro and other symbols
\else % if luatex or xetex
  \usepackage{unicode-math}
  \defaultfontfeatures{Scale=MatchLowercase}
  \defaultfontfeatures[\rmfamily]{Ligatures=TeX,Scale=1}
\fi

\ifpdf
  \DeclareUnicodeCharacter{2610}{%
    \makebox[10pt][l]{$\square$}\raisebox{.15ex}{\hspace{0.1em}\vphantom{$\checkmark$}}%
  }
  \DeclareUnicodeCharacter{2611}{%
    \makebox[0pt][l]{$\square$}\raisebox{.15ex}{\hspace{0.1em}$\checkmark$}%
  }
  \iftrue
    \usepackage{tikz}
    \newcommand*\circledtext[1]{\tikz[baseline=(char.base)]{%
        \node[shape=circle,draw,inner sep=0pt] (char) {#1};%
      }%
    }
  \fi

  \DeclareUnicodeCharacter{2460}{\circledtext{1}}
  \DeclareUnicodeCharacter{2461}{\circledtext{2}}
  \DeclareUnicodeCharacter{2462}{\circledtext{3}}
  \DeclareUnicodeCharacter{2463}{\circledtext{4}}
  \DeclareUnicodeCharacter{2464}{\circledtext{5}}
  \DeclareUnicodeCharacter{2465}{\circledtext{6}}
  \DeclareUnicodeCharacter{2466}{\circledtext{7}}
  \DeclareUnicodeCharacter{2467}{\circledtext{8}}
  \DeclareUnicodeCharacter{2468}{\circledtext{9}}

\fi

\usepackage{xcolor}
\usepackage{graphicx}

\usepackage[normalem]{ulem}

\usepackage{hyperref}
\usepackage{xurl} % add URL line breaks if available
\usepackage{listings}

\usepackage{booktabs} % For \toprule, \midrule and \bottomrule
\usepackage{datatool}

\colorlet{punct}{red!60!black}
\definecolor{background}{HTML}{EEEEEE}
\definecolor{delim}{RGB}{20,105,176}
\colorlet{numb}{magenta!60!black}

\lstdefinelanguage{json}{
    basicstyle=\normalfont\ttfamily,
    numbers=left,
    numberstyle=\scriptsize,
    stepnumber=1,
    numbersep=8pt,
    showstringspaces=false,
    breaklines=true,
    frame=lines,
    backgroundcolor=\color{background},
    literate=
     *{0}{{{\color{numb}0}}}{1}
      {1}{{{\color{numb}1}}}{1}
      {2}{{{\color{numb}2}}}{1}
      {3}{{{\color{numb}3}}}{1}
      {4}{{{\color{numb}4}}}{1}
      {5}{{{\color{numb}5}}}{1}
      {6}{{{\color{numb}6}}}{1}
      {7}{{{\color{numb}7}}}{1}
      {8}{{{\color{numb}8}}}{1}
      {9}{{{\color{numb}9}}}{1}
      {:}{{{\color{punct}{:}}}}{1}
      {,}{{{\color{punct}{,}}}}{1}
      {\{}{{{\color{delim}{\{}}}}{1}
      {\}}{{{\color{delim}{\}}}}}{1}
      {[}{{{\color{delim}{[}}}}{1}
      {]}{{{\color{delim}{]}}}}{1},
}

\hypersetup{
  pdftitle={Large-scale data extraction from the UNOS organ donor documents}
}

\title{Large-Scale Data Extraction From the OPTN Organ Donor Documents}

\author[1]{Marek Rychlik}
\affil[1]{Department of Mathematics, University of Arizona, Tucson, AZ 85721}
\author[2]{Bekir Tanriover}
\affil[2]{Department of Medicine, University of Arizona, Tucson, AZ 85721}
\author[3]{Yan Han}
\affil[3]{Research Engagement, The University of Arizona Libraries, Tucson, AZ 85721}
\graphicspath{{images/}}

\begin{document}
\maketitle

\begin{abstract}

  In this paper we focus on three major task: 1) discussing our
  methods: Our method captures a portion of the data in DCD
  flowsheets, kidney perfusion data, and Flowsheet data captured
  peri-organ recovery surgery. 2) demonstrating the result: We built a
  comprehensive, analyzable database from 2022 OPTN data. This dataset
  is by far larger than any previously available even in this
  preliminary phase; and 3) proving that our methods can be extended
  to all the past OPTN data and future data.

  The scope of our study is all Organ Procurement and Transplantation
  Network (OPTN) data of the USA organ donors since 2008.  The data
  was not analyzable in a large scale in the past because it was
  captured in PDF documents known as ``Attachments'', whereby every
  donor's information was recorded into dozens of PDF documents in
  heterogeneous formats. To make the data analyzable, one needs to
  convert the content inside these PDFs to an analyzable data format,
  such as a standard SQL database. In this paper we will focus on 2022
  OPTN data, which consists of $\approx 400,000$ PDF documents
  spanning millions of pages. The entire OPTN data covers 15 years
  (2008--20022). This paper assumes that readers are familiar with the
  content of the OPTN data.

\end{abstract}

\section{Background}
We introduce a multi-faceted, interdisciplinary problem with paramount
importance to millions of people worldwide, and especially to $\approx 550,000$
Americans with end-stage kidney disease (ESKD) who require
dialysis. We aim to address the inefficiencies in organ quality
assessment by making previously inaccessible data analyzable through
large-scale curation. The symptom of under-par performance of the
system is gradual increase in the number of organs that are not used
(discarded), e.g., nearly 30\% of recovered kidneys are not
transplanted, gradually increasing from 8\% in the 1980s. If the
non-use ratio is reduced, potentially thousands of lives would be saved
in the U.S. alone, and dramatically improve life quality for most
patients who spent 12 hours a day undergoing hemodialysis which
results from a successful kidney transplant. The total cost of
dialysis in the US is over \$30 billion (\$100,000 per patient per
year) and simple cost analysis shows that a kidney transplant reduces
this to a small fraction, potentially a 3-fold cost reduction per
patient.  The interdisciplinary research and software here described
in this paper target the inefficiencies flow of information in the
U.S. organ transplant system, where the primary vehicle for
communications is a collection of PDF documents (``PDF attachments'')
associated with every donated organ.

The interdisciplinary research and software here described in this
paper target the inefficiencies flow of information in the US organ
transplant system, where the primary vehicle for communications is a
collection of PDF documents (``PDF attachments'') associated with
every donated organ. The subject of investigation by our
interdisciplinary group is the record of all US organ transplants
since electronic data collection started (2008-2022) acquired from
OPTN. For practical reasons in this stage of our project we narrowed
the dataset to the 2022 portion (10\% of the data), which is
representative of the entire data, but much easier to manage in the
software development cycle. In the current paper we report
the preliminary results obtained from this portion of the data,
with future publications planned to cover all aspects of the project.

\label{sec:background}
\subsection{UNOS, Organ Procurement Organizations and DonorNet}
\label{unos-organ-procurment-organizations-and-donornet}

In 1984 the United States 98th Congress passed the National Organ
Transplant Act (NOTA: P.L. 98-507) to regulate organ donation,
matching, and placement process \cite{Public_Law_98_507}. The NOTA
launched the Organ Procurement and Transplantation Network (OPTN)
which has been operated by a private, independent, and non-profit
organization called United Network for Organ Sharing (UNOS) under the
supervision of the Federal government during the past four decades
(\href{https://unos.org/about/national-organ-transplant-system/}{https://unos.org/about/national-organ-transplant-system/}). UNOS
develops organ allocation policies, manages the national transplant
waitlist, matches donors to recipients via a computerized algorithm,
maintains an organ transplant database, and educates the public and
transplant community.

When a patient is diagnosed with brain death while being treated as an
inpatient, U.S. donor hospitals are obligated to inform and
collaborate with Organ Procurement Organizations (OPO) operating
within their respective donor service area (DSA). The OPOs, which are
not-for-profit entities, are regulated by the Centers for Medicare and
Medicaid Services (CMS). The United States has a total of 57 OPOs,
each holds the responsibility of obtaining consent from donor
families, transmitting donor medical data to UNOS via UNet (a web
portal that connects the OPOs and transplant centers), procuring
organs, and facilitating their delivery to over 250 transplant centers
(TC). The OPTN uses UNet as a main tool to collect and verify data,
which was implemented on October 25, 1999
\cite{Gerber_Arrington_Taranto_Baker_Sung_2010,
  Massie_Zeger_Montgomery_Segev_2009}. The OPOs have access to donor
data at more than 1,000 donor hospitals in real-time. UNOS completed
integration of DonorNet into UNet in 2008, which enables the OPOs to
upload and update deceased donor information digitally.

\subsection{Source Data and PDF Attachments}
\label{source-data-and-pdf-attachments}

Depending on the OPOs, the forms/content used to upload the same donor
information may vary. For example, there are over 25 different types
of forms used for kidney anatomy and pathology reporting. By reviewing
critical information in the donor attachments, the TC of receiving
organ offers can make informed decisions regarding organ utilization
and the system may improve efficiency of organ placement
\cite{Massie_Zeger_Montgomery_Segev_2009}. The TCs build their
provisional acceptance decision of a kidney offer predominantly based
on donors' characteristics, which is summarized in Kidney Donor Profile Index (KDPI). At the time of a
deceased donor organ recovery surgery, the surgical team generally
communicates donor anatomy (kidney length, vascular plaques in aorta
and renal artery, number of renal arteries and ureters, kidney tumors,
infarcts, thrombosis) to the accepting TC. Organ utilization decision
is mostly finalized based on a combination of post-recovery data (e.g.
donor anatomy \cite{Tierie_Roodnat_Dor_2019,
  Keijbeck_Veenstra_Pol_Konijn_Jansen_vanGoor_Hoitsma_Peutz-Kootstra_Moers_2020,
  Woestenburg_Sennesael_Bosmans_Verbeelen_2008,
  Husain_King_Robbins-Juarez_Adler_McCune_Mohan_2021,
  Dare_Pettigrew_Saeb-Parsy_2014} , recovery type and associated cold
and warm ischemia time
\cite{Gill_Rose_Lesage_Joffres_Gill_OConnor_2017,
  Brennan_Sandoval_Husain_King_Dube_Tsapepas_Mohan_Ratner_2020},
procurement kidney biopsy results
\cite{Wang_Wetmore_Crary_Kasiske_2015, Lentine_Kasiske_Axelrod_2021,
  Husain_Shah_AlvaradoVerduzco_King_Brennan_Batal_Coley_Hall_Stokes_Dube_etal._2020,
  Carpenter_Husain_Brennan_Batal_Hall_Santoriello_Rosen_Crew_Campenot_Dube_etal._2018,
  Lentine_Naik_Schnitzler_Randall_Wellen_Kasiske_Marklin_Brockmeier_Cooper_Xiao_etal._2019},
and pulsatile machine perfusion values
\cite{Tingle_Figueiredo_Moir_Goodfellow_Talbot_Wilson_Tingle_2019}.

\begin{comment}
\sout{The effects of post-recovery data abnormalities on deceased
  donor kidney utilization (accept vs reject) and transplant
  outcomes are understudied and those studies suffer from small
  sample sizes (N\textless6,000)
  \cite{Stewart_Foutz_Kamal_Weiss_McGehee_Cooper_Gupta_2022}. Additionally,
  heterogeneity and uncertainty in offer acceptance decision making
  increases complexity.}
\end{comment}

Over 99\% of donor data (including detailed post-recovery data) has been
stored in each donor's profile as individual PDF attachments in the
DonorNet since 2008. There are 135,446 deceased
donors with over 1 million PDF attachments between 2008 to 2021
\cite{The_UNOS_STAR_Deceased_Donor_SAF_file}, as each deceased donor
in the DonorNet generally has 10--20 PDF attachments. \yan{Note: This
  number does not add up : 10 x 135,000 =1.35 million } These PDF attachments consist of donor forms (information captured in
checkboxes, handwritten digits, and comments, typed text and/or
handwritten annotations), image-only content such as photographs taken
during organ procurement, X-rays, and digital pathology slides.
Furthermore, short video clips featuring echocardiograms, angiograms,
and bronchoscopies may be attached.  A detailed discussion of the content of the PDF attachments and challenges that they pose is available in the other paper. \cite{article1-2023}

\subsection{The focus of the preliminary study }
The collection of 1 million PDF attachments represents a substantial
volume of data, totaling approximately 10 terabytes in size. Given
that the data consisting of individual medical records and health
information, it is required to process the data within a
HIPAA-compliant environment. This makes this problem formidable in
terms of computing resources and organizational effort required to
perform the tasks described in this paper. We are currently using the
new HIPAA-compliant HPC cluster at the University of Arizona called
\href{https://soteria.arizona.edu/}{Soteria}.

In this phase we focused on data in native-born (``pure'') PDFs. While
PDF files are primarily designed for storing and presenting textual
and graphical content, they can also include binary data such as
images, audio, video, and other non-textual elements. In other words,
PDF is a flexible container file format.

This distinction is important because handling documents stored as
images requires additional processing, including Optical Character
Recognition (OCR). Moreover, many images are captured by smartphones'
cameras, and are subject to perspective distortion, which requires
application of computer vision techniques.  We have developed relevant
techniques over the past few years, but building a complete set of
software tool kits capable of processing the entire OPTN dataset has
not been completed because of lack of sufficient funding.

The current study was approved by the University of Arizona Institutional Review Board (IRB).

\section{A case study: extraction of DCD flowsheets}
\begin{figure}
  \marek{To do subcaption box style split figures look at this:
    \url{http://mirrors.ibiblio.org/CTAN/macros/latex/contrib/caption/subcaption.pdf}}
  \subcaptionbox{DCD flowsheet --- page~1\label{fig:dcd-flowsheet-p1}}
  {\includegraphics[width=0.45\textwidth]{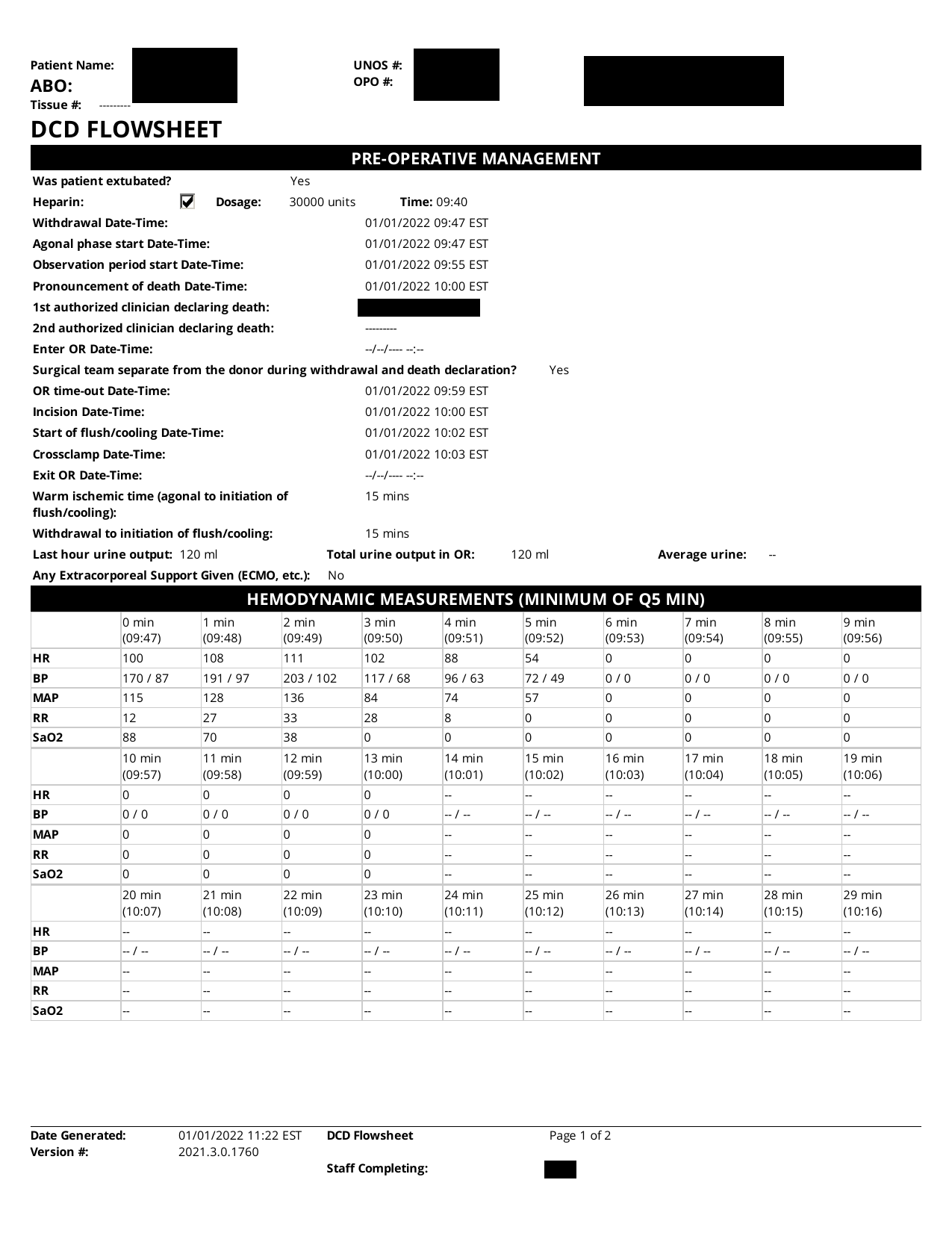}}
  \subcaptionbox{DCD flowsheet --- page~2\label{fig:dcd-flowsheet-p2}}
  {\includegraphics[width=0.45\textwidth]{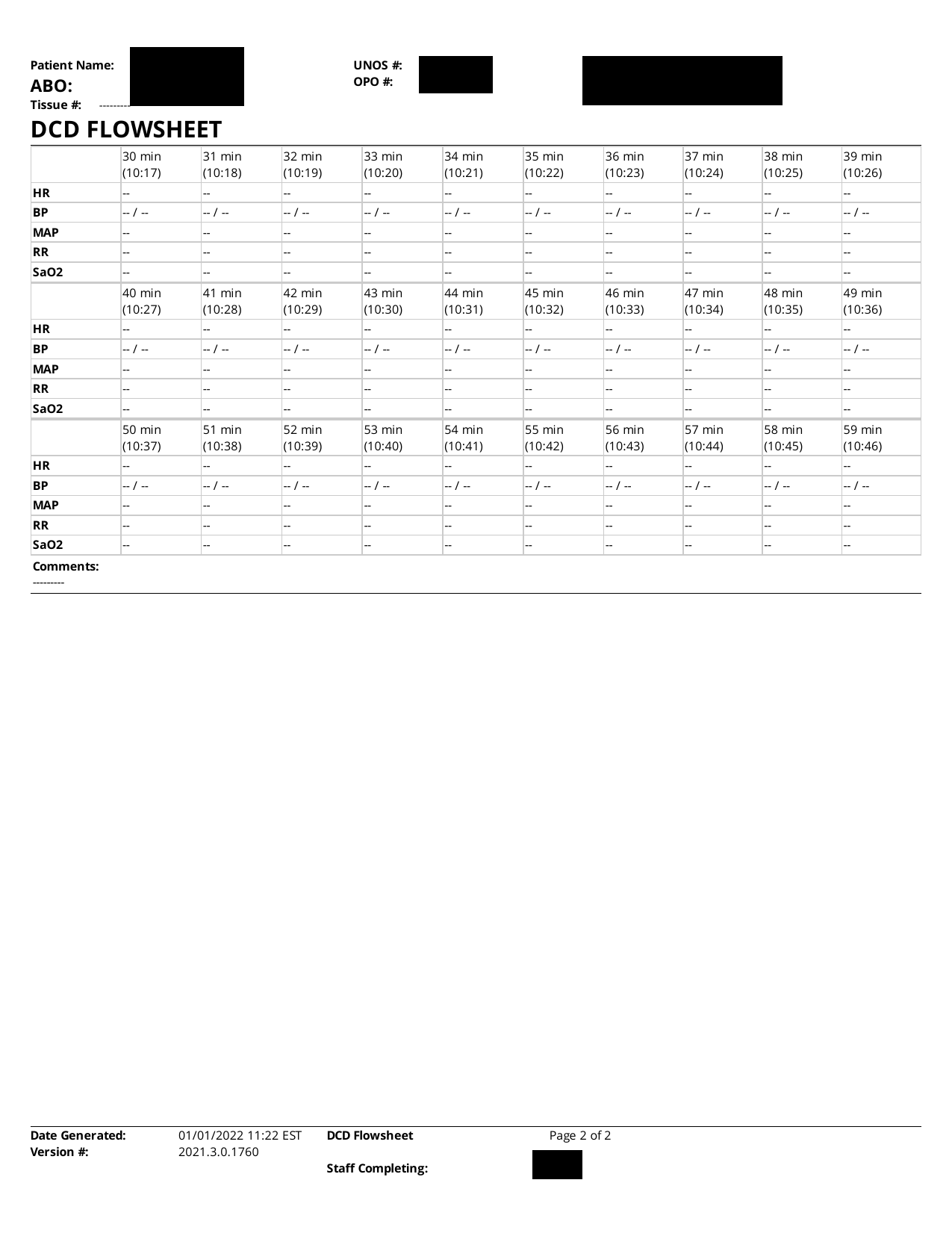}}
  \caption{A DCD FLowsheet Variant (PII redacted).\label{fig:dcd-flowsheet}}
\end{figure}

The PDF attachments of interest in the case of DCD donors include the
\emph{DCD Flowsheet} shown in Figure~\ref{fig:dcd-flowsheet}. The main
data in this form is a time series representing vital signs of a donor
in the agonal phase, to be used as predictor of the organ damage, and
ultimately determines the graft survival and durability.

Extraction of the time series from PDF documents consists of complex
tasks. The textual data can be extracted from PDF in a straightforward
fashion (without OCR), and is 100\% accurate (in the absence of
software glitches and human error).  For this task we use open
software written in Java, \emph{PDFBox}, along with an enhancement of
the PDF text extraction \emph{PDFLayoutTextStripper}. This software
extracts \emph{formatted text}. Unlike the software provided inside
\emph{PDFBox} called \emph{PDFTextStripper}, the enhanced version
preserves the layout of the text on the page, thus allowing us to
sufficiently correct page segmentation (e.g., discovery of table
boundaries).  It should be mentioned that PDF is a rich format, which
allows storing data tables in native PDF tables. However, these
mechanisms are not used in the PDF attachments. The reason can be
traced to the fact that the original document is in the
\emph{Microsoft Word} format (.docx) and the PDF is a product of
exporting the document to PDF file format. Multiple tools are used to
perform the Word-to-PDF conversion, resulting in PDFs that differ
internally and significantly enough to trip up schemes relying upon
cutting-edge features of the PDF format. The approximation of the
plain text content of the PDF preserving layout turns out to be the
lowest common denominator which allows achieving almost all objectives
related to data extraction.

Other data, besides the time series, includes a survey with
$\approx 20$ questions with answers which are either categorical
(often: Yes/No) or factual questions such as dates, person names, and
volumes. The data is relatively easy to extract. The main issue is
that the lack of standard formatting and many ways to refer to the
same categorical variable value.

\subsection{The ``Checkbox~Problem'' and our solution}
One critical problem is to decode information captured by checkbox questions. 
Some answers are provided as checkmarks (checked or unchecked box).
For example, in the 'DCD flowsheet'
(Figure~\ref{fig:dcd-flowsheet-p1}) the question 'Heparin:' (i.e.,
whether heparin was used) is answered by checking a box. While the
field 'Dosage:' and 'Time:' are indications that heparin was used,
allowing to infer the answer the question 'Heparin' indirectly in this
form, other forms make much more extensive use of checkboxes, with
some forms including $>100$ checkboxes on a single page. Therefore, the
``Checkbox Problem'' needed to be solved, and the effort spent was
justified.

Ideally, the checked and unchecked box would be a part of the PDF
text, as they are valid Unicode characters (Unicode characters 2610
and 2611). Unfortunately, the great majority of PDF forms containing
checkboxes do not use the designated Unicode for the purpose. Instead, the checkbox
is drawn by PDF drawing commands (instructions that draw parts of the
box as lines, and shading portions of the box).  The important
observation is that \textbf{PDF standards include a programming language} allowing
inclusion of programs with graphical output, such as the checkboxes.
However, it is a very wasteful way to store binary information, such
as 1 bit of data telling us whether heparin was used or not.  A
detailed study of the PDF in question revealed that the particular
checkmark in the DCD~Flowsheet form is stored in a program that
requires 18,000 bits to store. Furthermore, it is virtually impossible
to identify a large quantity of checkmarks by looking at the PDF as a
computer program.

The approach we developed uses features of PDF and their implementations
in \emph{PDFBox} to our advantage. The PDF is viewed as a stream of
objects, some representing textual information and some representing
graphics. The two types of objects can be separated by writing a Java
program using the \emph{PDFBox} API.  We used this capability of
\emph{PDFBox} to extract all textual information.  The remaining
graphical objects were stored in a new PDF document, which was
subsequently converted to JPEG, and subjected to image processing.
\begin{figure}
  \marek{To do subcaption box style split figures look at this:
    \url{http://mirrors.ibiblio.org/CTAN/macros/latex/contrib/caption/subcaption.pdf}}
  \subcaptionbox{DCD flowsheet --- page~1, graphical (non-text) objects.\label{fig:dcd-flowsheet-p1-graphics}}
  {\includegraphics[width=0.45\textwidth]{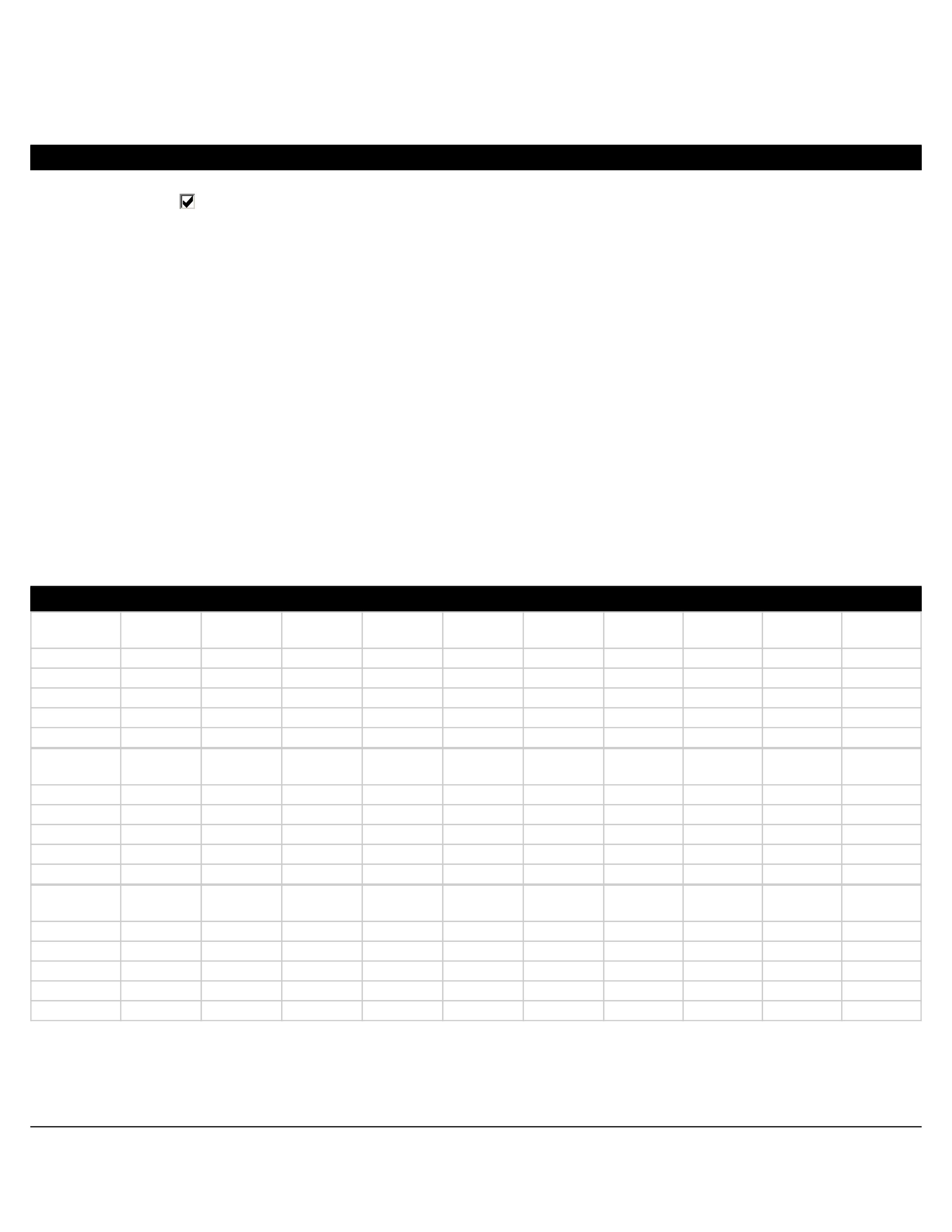}}
  \subcaptionbox{DCD flowsheet --- page~1, graphical (non-text) objects.\label{fig:dcd-flowsheet-p2-graphics}}
  {\includegraphics[width=0.45\textwidth]{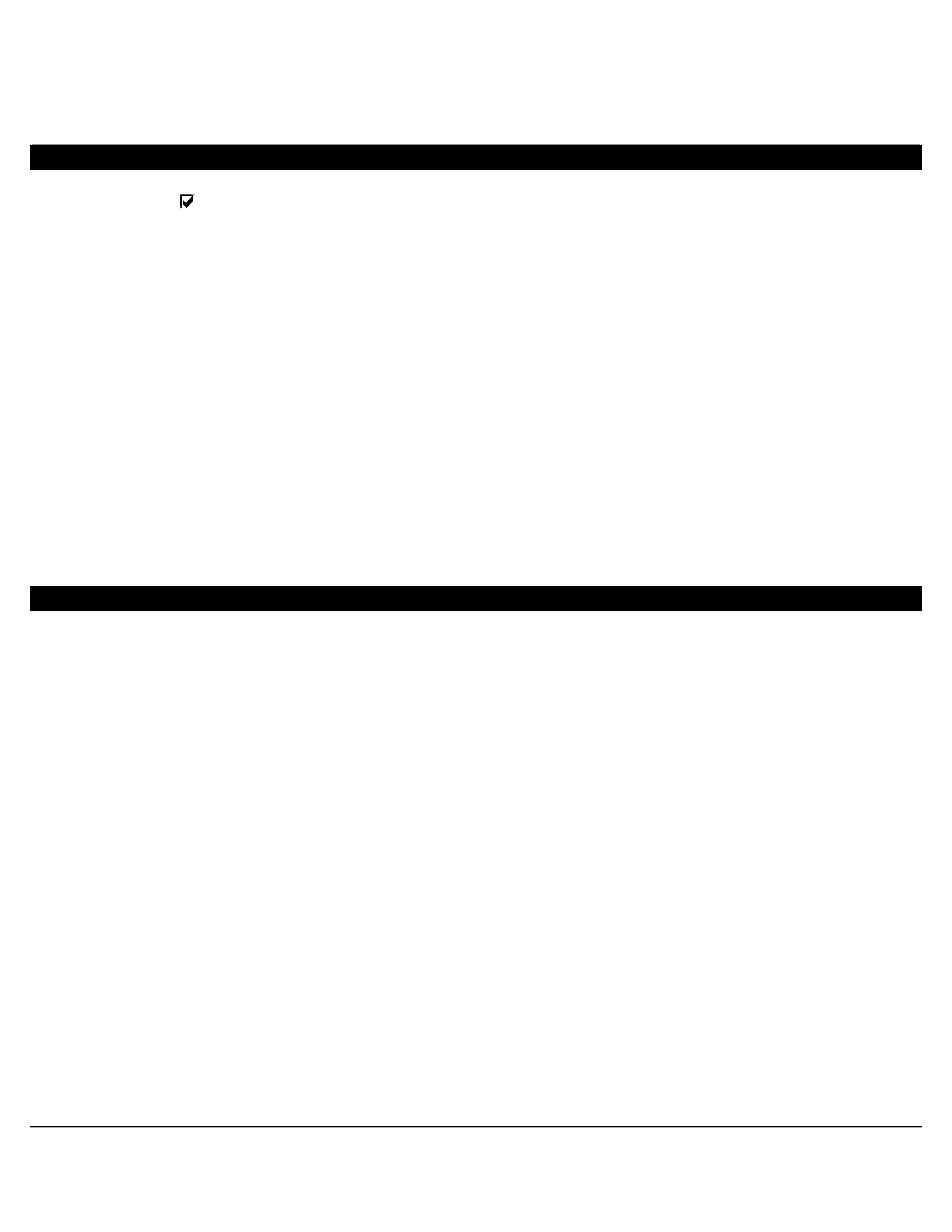}}
  \caption{DCD Flowsheet --- graphical objects and binarization.}\label{fig:dcd-flowsheet-graphics}
\end{figure}
The tools were provided by commercial software MATLAB, which is the
main framework used in this project. MATLAB provides extensions known
as ``toolboxes'' and the ``Image Processing Toolbox'' has been the
tool of choice for decades for tasks such as image segmentation.  The
ability to detect and classify ``objects'' in images based on diverse
criteria is provided by the functions \emph{bwconn} (detector of
connected components in monochrome images), \emph{regionprops} (an
extensive object feature extractor). We suppressed the table
boundaries based on the fact that they are gray rather than black, by
using command \emph{imbinarize}. While this command is very useful in
analyzing OCR-based documents, in our case we can just use
thresholding to eliminate table boundaries. Finally, the checked and
unchecked boxes can be easily distinguished, by setting a threshold
for the number of foreground pixels (obviously, checked boxes have
many more foreground pixels). Figure~\ref{fig:dcd-flowsheet-graphics}
shows the image with all graphical objects, and the result
of its binarization.

\section{Regular languages, grammars and regular expressions}
The PDF attachments and medical forms in general present an interesting case of
a restricted domain language. Natural language processing (NLP) and computer science (theory
of compilation) has made an extensive use of \emph{regular grammars}, a relatively
simple mechanisms for capturing syntax of relatively simple languages. Our
method of parsing the documents in question is based on the premise that
medical forms can be accurately parsed by regular grammars.

In large problems, such as ours, we are dealing with dozens of medical
form types, with many variations amongst them. Therefore, it is
important to streamline the process of constructing form-specific
regular grammars. In our work to date we constructed them mostly ``by-hand''.
A rough algorithm is this:
\begin{enumerate}
\item Extract plain text of a page of a document
\item Note fixed text (queries, table headings, etc.)
\item Note the data type of the variable portions of the
  form and construct a regular expression that captures
  their structure best. This includes numerical data,
  dates, names, etc.
\item Try parsing a representative sample of documents
  in question
\item Identify failures, repeat.
\end{enumerate}
Typically, after several iterations of this process most documents
($>90\%$) are matched and parsed correctly. The byproduct of parsing
is assigning semantic values to the patterns representing the variable
portions of medical forms, i.e. user input.  The process of mapping
the obtained values to a database is relatively straightforward.

There are two main ways to represent regular grammars:
\begin{enumerate}
\item use BNF notation;
\item use regular expression.
\end{enumerate}
In applications like ours, the approach based on constructing regular
expression appears more productive. As an example,
the regular expression in Table~\ref{tbl:preop-section} matches the text
at top of the first page of the DCD flowsheet (``PRE-OPERATIVE~MANAGMENT'' section).
Reading and understanding this page-long regular expression is a daunting task.
It should be noted that there are two possible approaches:
\begin{enumerate}
\item Top down, as described above, where variable portions of the
  text are replaced with small regular expressions.
\item Bottom up approach, where regular expressions are constructed
  for smaller portions of the text, such as lines of text, and
  combined
\end{enumerate}
We found the bottom-up approach quicker when building the expressions
``by hand''.  The term ``by hand'' does not accurately reflect the
process, as in practice we developed a computer program that writes the
regular expression in Table~\ref{tbl:preop-section}.

\begin{table}
  \caption{The regular expression used in parsing of the
    ``PRE-OPERATIVE~MANAGEMENT'' section of the first page of the
    DCD~Flowsheet (PII redacted).\label{tbl:preop-section}}
\begin{lstlisting}[language=Perl,
  basicstyle=\tiny\ttfamily,
  columns=flexible,
  breaklines=true
  ]
[ ]*(?<extubated_key>Was patient extubated\?)[
]+(?<extubated>[a-zA-Z-]+)[ ]*\n[ ]+(?<heparin_key>Heparin:)[
]*(?<heparin_dosage_key>Dosage:)?[
]*(?<heparin_dosage>(?(heparin_dosage_key)\d+[ ]+units)|)[
]*(?<heparin_time_key>Time:)?[
]*(?<heparin_time>(?(heparin_time_key)((\d+):(\d+))|))?[ ]*\n[
]*(?<witdrawal_datetime_key>Withdrawal Date-Time:)[
]+(?<withdrawal_datetime>[0-9-]{2,2}/[0-9-]{2,2}/[0-9-]{4,4}[
]+[0-9-]{2,2}:[0-9-]{2,2})[
]+(?<withdrawal_datetime_tz>(MST|EST|PDT|CST|EDT|PST|CDT|MDT|AST|HST))?[
]*\n[ ]*(?<agonal_phase_start_datetime_key>Agonal phase start[
]+Date-Time):[
]+(?<agonal_phase_start_datetime>[0-9-]{2,2}/[0-9-]{2,2}/[0-9-]{4,4}[
]+[0-9-]{2,2}:[0-9-]{2,2})[
]+(?<agonal_phase_start_datetime_tz>(MST|EST|PDT|CST|EDT|PST|CDT|MDT|AST|HST))?[
]*\n[ ]*(?<observation_period_start_datetime_key>Observation period
start[ ]+Date-Time):[
]+(?<observation_period_start_datetime>[0-9-]{2,2}/[0-9-]{2,2}/[0-9-]{4,4}[
]+[0-9-]{2,2}:[0-9-]{2,2})[
]+(?<observation_period_start_datetime_tz>(MST|EST|PDT|CST|EDT|PST|CDT|MDT|AST|HST))?[
]*\n[ ]*(?<pronouncement_of_death_datetime_key>Pronouncement[ ]+of[
]+death[ ]+Date-Time):[
]+(?<pronouncement_of_death_datetime>[0-9-]{2,2}/[0-9-]{2,2}/[0-9-]{4,4}[
]+[0-9-]{2,2}:[0-9-]{2,2})[
]+(?<pronouncement_of_death_datetime_tz>(MST|EST|PDT|CST|EDT|PST|CDT|MDT|AST|HST))?[
]*\n[ ]+(?<first_authorized_clinician_declaring_death_key>1st
authorized clinician declaring death):[ ]+(?<doctor_name>([a-zA-Z.,
]+[a-zA-Z]|[-]+))[ ]*\n[
]+(?<second_authorized_clinician_declaring_death_key>2nd[
]+authorized[ ]+clinician[ ]+declaring[ ]+death):[
]+((?<doctor_name>([a-zA-Z., ]+[a-zA-Z]|[-]+))|n/a)[ ]*\n[
]*(?<enter_or_datetime_key>Enter OR Date-Time):[
]+(?<enter_or_datetime>[0-9-]{2,2}/[0-9-]{2,2}/[0-9-]{4,4}[
]+[0-9-]{2,2}:[0-9-]{2,2})[
]+(?<enter_or_datetime_tz>(MST|EST|PDT|CST|EDT|PST|CDT|MDT|AST|HST))?[
]*\n[ ]+(?<surgical_team_separate_key>Surgical[ ]+team[ ]+separate[
]+from[ ]+the donor[ ]+during[ ]+withdrawal[ ]+and[ ]+death[
]+declaration\?)[ ]*(?<surgical_team_separate>(Yes|No|[-]+))[ ]*\n[
]*(?<or_time_out_datetime_key>OR time-out Date-Time):[
]+(?<or_time_out_datetime>[0-9-]{2,2}/[0-9-]{2,2}/[0-9-]{4,4}[
]+[0-9-]{2,2}:[0-9-]{2,2})[
]+(?<or_time_out_datetime_tz>(MST|EST|PDT|CST|EDT|PST|CDT|MDT|AST|HST))?[
]*\n[ ]*(?<incision_datetime_key>Incision[ ]+Date-Time):[
]+(?<incision_datetime>[0-9-]{2,2}/[0-9-]{2,2}/[0-9-]{4,4}[
]+[0-9-]{2,2}:[0-9-]{2,2})[
]+(?<incision_datetime_tz>(MST|EST|PDT|CST|EDT|PST|CDT|MDT|AST|HST))?[
]*\n[ ]*(?<start_of_flush_cooling_datetime_key>Start of flush/cooling[
]+Date-Time):[
]+(?<start_of_flush_cooling_datetime>[0-9-]{2,2}/[0-9-]{2,2}/[0-9-]{4,4}[
]+[0-9-]{2,2}:[0-9-]{2,2})[
]+(?<start_of_flush_cooling_datetime_tz>(MST|EST|PDT|CST|EDT|PST|CDT|MDT|AST|HST))?[
]*\n[ ]*(?<crossclamp_datetime_key>Crossclamp[ ]+Date-Time):[
]+(?<crossclamp_datetime>[0-9-]{2,2}/[0-9-]{2,2}/[0-9-]{4,4}[
]+[0-9-]{2,2}:[0-9-]{2,2})[
]+(?<crossclamp_datetime_tz>(MST|EST|PDT|CST|EDT|PST|CDT|MDT|AST|HST))?[
]*\n[ ]*(?<exit_or_datetime_key>Exit[ ]+OR[ ]+Date-Time):[
]+(?<exit_or_datetime>[0-9-]{2,2}/[0-9-]{2,2}/[0-9-]{4,4}[
]+[0-9-]{2,2}:[0-9-]{2,2})[
]+(?<exit_or_datetime_tz>(MST|EST|PDT|CST|EDT|PST|CDT|MDT|AST|HST))?[
]*\n[ ]+(?<warm_ischemic_time_key>Warm ischemic time) \(agonal to
initiation of[ ]+(?<minutes>(\d+|[-]+))[ ]+mins[ ]*\n[
]+flush/cooling\):[ ]*\n[
]+(?<withdrawal_to_initiation_of_flush_cooling_key>Withdrawal to
initiation of flush/cooling):[ ]+ (?<minutes>(\d+|[-]+))[ ]+mins[
]*\n[ ]+(?<last_hour_urine_output_key>Last hour urine output):
(?<last_hour_urine_output>[0-9-]+)[ ]+ml[
]+(?<total_urine_output_in_or_key>Total urine output in OR):
(?<total_urine_output_in_or>[0-9.-]+)[ ]+ml[
]+(?<average_urine_key>Average urine:)[
]+(?<average_urine>([0-9.-]+)?[ ]*(ml/hr)?)[ ]*\n[
]+(?<extracorporeal_support_key>Any Extracorporeal Support Given
\(ECMO, etc\.\)):[ ]+(?<extracorporeal_support>(Yes|No|[-]+))[
]*(?<how_long_key>How[ ]+Long:)?[ ]*(?<how_long>(?(how_long_key)\d+[
]+hrs|))[ ]*(?<flow_rate_key>Flow[ ]+Rate:)?[
]*(?<flow_rate>(?(flow_rate_key)\d+(\.\d+)?[ ]+L/min|))[ ]*\n
\end{lstlisting}
\end{table}

\begin{figure}[htb]
  \marek{To do subcaption box style split figures look at this:
    \url{http://mirrors.ibiblio.org/CTAN/macros/latex/contrib/caption/subcaption.pdf}}
  \subcaptionbox{The LIVER~DATA form.\label{fig:liver-data-original}}
  {\includegraphics[width=0.45\textwidth]{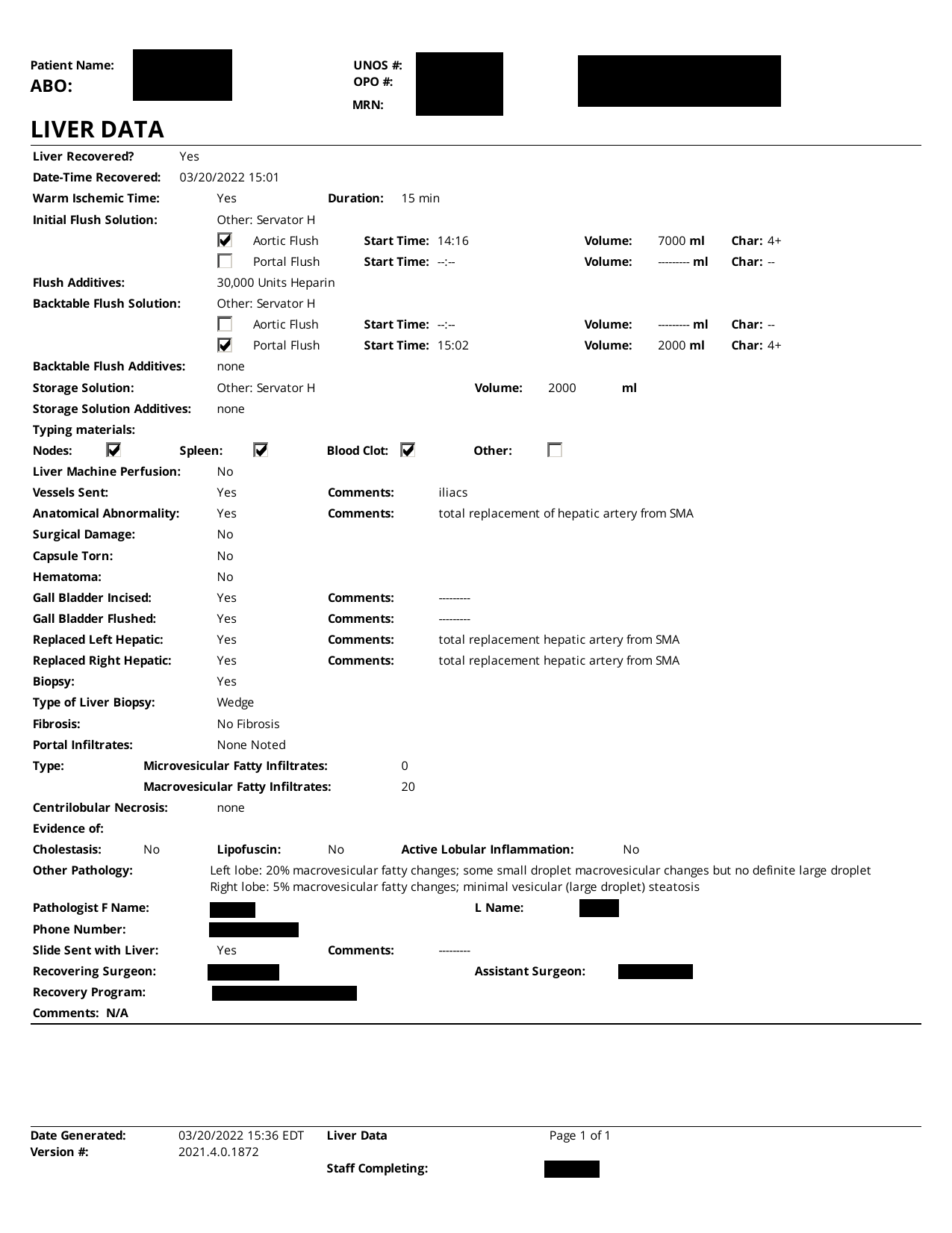}}
  \subcaptionbox{The graphical objects of the form.\label{fig:liver-data-binarized}}
  {\includegraphics[width=0.45\textwidth]{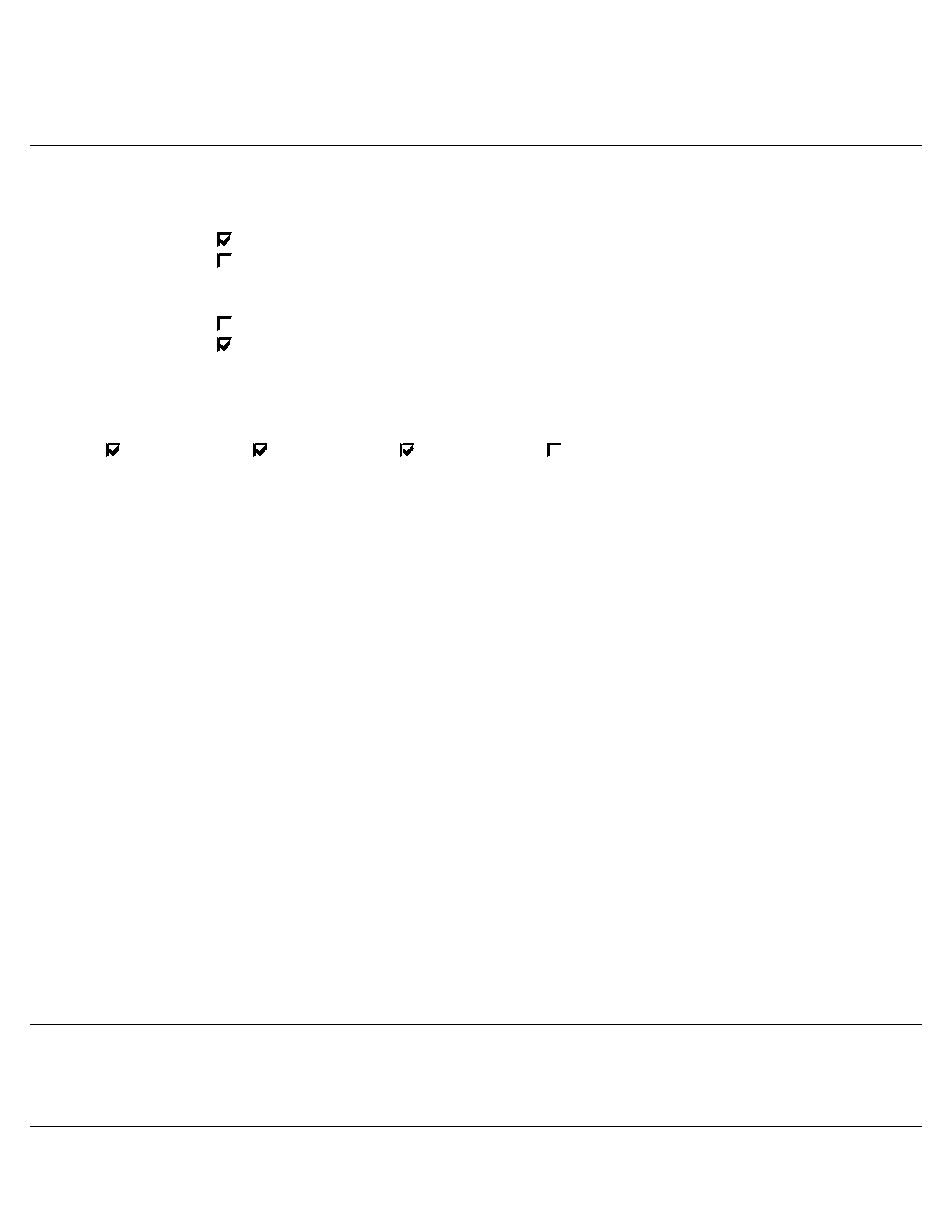}}
  
  \subcaptionbox{The eight checkboxes.\label{fig:liver-data-8-checkboxes}}
  {\centering\includegraphics[width=0.3\textwidth]{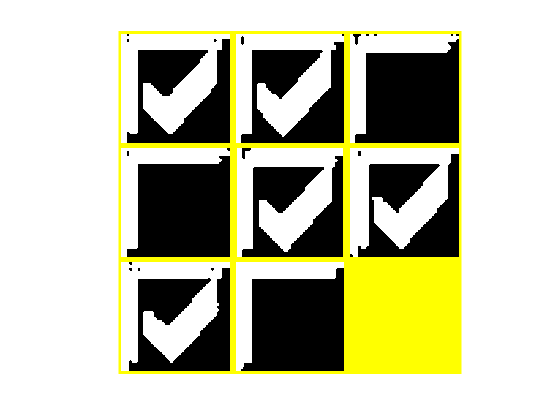}}
  \caption{The LIVER~DATA form --- graphical objects and binarization  (PII redacted).\label{fig:liver-data}}
\end{figure}

The regular expression, such as the one in
Table~\ref{tbl:preop-section} is of tremendous value because it is a
concise and readable (with some practice), description of a language
parser. This description allows to complete the task of parsing with
one line of code using MATLAB (and other high-level software
environments). E.g., in MATLAB the code is this
\begin{lstlisting}[language=MATLAB]
  names = regexp(page_text, pattern, 'names');
\end{lstlisting}
The system builds a data structure which binds named tokens, such as
'(?<extubated>)', '(?<heparine\_dosage>)', etc., to their semantic values
set by user when filling out the form, located at the lines 3 and 4 in
the JSON output of Table~\ref{tbl:preop-section-json}.

\begin{table}
  \caption{The content of the ``PRE-OPERATIVE~MANAGEMENT'' section of the DCD Flowsheet
    captured as JSON, equivalent to a database or spreadsheet table. JSON was chosen
    because the equivalent spreadsheet table would not fit due to its large width.\label{tbl:preop-section-json}}
\begin{lstlisting}[language=json,
    basicstyle=\small\ttfamily]
[
    {
        "extubated": "Yes",
        "heparin_dosage": 30000,
        "heparin_dosage_unit": "units",
        "heparin_time": "09:40",
        "withdrawal_datetime": "2022-01-01 09:47 EST",
        "withdrawal_datetime_tz": "EST",
        "agonal_phase_start_datetime": "2022-01-01 09:47 EST",
        "agonal_phase_start_datetime_tz": "EST",
        "observation_period_start_datetime": "2022-01-01 09:55 EST",
        "observation_period_start_datetime_tz": "EST",
        "pronouncement_of_death_datetime": "2022-01-01 10:00 EST",
        "pronouncement_of_death_datetime_tz": "EST",
        "doctor_name": "XXXXXXXXXXXXXXXXXXXXX",
        "enter_or_datetime": null,
        "enter_or_datetime_tz": "",
        "surgical_team_separate": "Yes",
        "or_time_out_datetime": "2022-01-01 09:59 EST",
        "or_time_out_datetime_tz": "EST",
        "incision_datetime": "2022-01-01 10:00 EST",
        "incision_datetime_tz": "EST",
        "start_of_flush_cooling_datetime": "2022-01-01 10:02 EST",
        "start_of_flush_cooling_datetime_tz": "EST",
        "crossclamp_datetime": "2022-01-01 10:03 EST",
        "crossclamp_datetime_tz": "EST",
        "exit_or_datetime": null,
        "exit_or_datetime_tz": "",
        "minutes": 15,
        "last_hour_urine_output": "120",
        "total_urine_output_in_or": "120",
        "average_urine": "--                            ",
        "extracorporeal_support": "No",
        "how_long": "",
        "flow_rate": "",
        "withdrawal_of_care_datetime": null,
        "withdrawal_of_care_datetime_tz": "",
        "withdrawal_of_care_location": "---------",
        "time_of_extubation_datetime": null,
        "time_of_extubation_datetime_tz": "",
        "physician_declaring_death": "---------",
        "person_notifying_family_of_death": "---------",
        "person_notifying_family_of_death_title": "---------",
        "recovery_team_enter_or_datetime": null,
        "recovery_team_enter_or_datetime_tz": "",
        "initiation_of_flush_cooling_datetime": null,
        "initiation_of_flush_cooling_datetime_tz": "",
        "minute_warm_ischemic_time_began": "---------",
        "warm_ischemic_time": "--"
    }
]
\end{lstlisting}
\end{table}

\section{A case study: extraction of liver data}
An example of a form entitled ``LIVER~DATA'' is shown in
Figure~\ref{fig:liver-data}.  As we can see, the form is similar to
the ``PRE-OPERATIVE~MANAGMENT'' section of the DCD~Flowsheet.  The
notable difference is the presence of 8 checkboxes. To correctly
decode this form we must decode the content of the checkboxes, as the
status of the checkmark cannot be inferred from the textual
information in the form. The method we outlined for the DCD~Flowsheets
works well here and the block of checkboxes can be seen in
Figure~\ref{fig:liver-data-8-checkboxes}. It should be noted that
MATLAB software scans objects left-to-right and then top-down,
resulting in the specific ordering of checkboxes. It should be noted
that although checkboxes in the PDF file are ``perfect'' and
resolution-independent in the PDF file, they must be rendered (in this
case, the resolution is 300dpi) and binarized, which results in minor
artifacts. However, all checkboxes are approximately 80-by-80 images,
and the threshold of 2500 foreground pixels separates the checked from
unchecked checkboxes well.

\section{Database Schema and preparing for data analysis}
Data is analyzable if it is amenable to statistical analysis.
The process of data preparation and normalization encompasses
\begin{enumerate}
\item Identification of captured variables.
\item Classification of variables into common classes: quantitative,
  ordered categorical, unordered categorical.
\end{enumerate}
In addition, to capture tens to hundreds of millions of records present in the dataset
we need to design a database schema which will permit efficient statistical
analysis and be suitable for machine learning.

It is straightforward to identify the variables, as they closely correspond to
the fields present in the forms. This leaves the problem of naming the variables
that promotes analysis and is compatible with other available databases.
An example of a database table with adopted variable naming conventions
is presented in Table~\ref{tbl:dcd_flowsheet_tbl}. The variable names
are chosen to work well as ``identifiers'' in most programming languages.
However, SQL allows any naming convention by using double quotes:
any string is a valid identifier (table column name) as long as it is quoted
with double quotes (").

We represented the time series data by a typical table which stores
time as one of the variable. It is noted that handling time in the PDF attachments
is far from perfect, because handling time correctly is a difficult
problem due to varying standards and daylight saving time. 

\def\NaN{{\tt NaN}}
\begin{table}
  \caption{DCD Flowsheet data table exported from MATLAB in CSV format.
    The data is truncated at 20 minutes, as all variables have
    missing values beyond this time point.\label{tbl:dcd_flowsheet_tbl}}
  \begin{tabular}{|c|c|c|c|c|c|c|c|}
    \hline
    Minute& Time& HR& BP\_Systolic& BP\_Diastolic& MAP& RR& SaO2\\
    \hline
    0& 2022-01-01 09:47 EST& 100& 170& 87& 115& 12& 88\\
    1& 2022-01-01 09:48 EST& 108& 191& 97& 128& 27& 70\\
    2& 2022-01-01 09:49 EST& 111& 203& 102& 136& 33& 38\\
    3& 2022-01-01 09:50 EST& 102& 117& 68& 84& 28& 0\\
    4& 2022-01-01 09:51 EST& 88& 96& 63& 74& 8& 0\\
    5& 2022-01-01 09:52 EST& 54& 72& 49& 57& 0& 0\\
    6& 2022-01-01 09:53 EST& 0& 0& 0& 0& 0& 0\\
    7& 2022-01-01 09:54 EST& 0& 0& 0& 0& 0& 0\\
    8& 2022-01-01 09:55 EST& 0& 0& 0& 0& 0& 0\\
    9& 2022-01-01 09:56 EST& 0& 0& 0& 0& 0& 0\\
    10& 2022-01-01 09:57 EST& 0& 0& 0& 0& 0& 0\\
    11& 2022-01-01 09:58 EST& 0& 0& 0& 0& 0& 0\\
    12& 2022-01-01 09:59 EST& 0& 0& 0& 0& 0& 0\\
    13& 2022-01-01 10:00 EST& 0& 0& 0& 0& 0& 0\\
    14& 2022-01-01 10:01 EST& \NaN& \NaN& \NaN& \NaN& \NaN& \NaN\\
    15& 2022-01-01 10:02 EST& \NaN& \NaN& \NaN& \NaN& \NaN& \NaN\\
    16& 2022-01-01 10:03 EST& \NaN& \NaN& \NaN& \NaN& \NaN& \NaN\\
    17& 2022-01-01 10:04 EST& \NaN& \NaN& \NaN& \NaN& \NaN& \NaN\\
    18& 2022-01-01 10:05 EST& \NaN& \NaN& \NaN& \NaN& \NaN& \NaN\\
    19& 2022-01-01 10:06 EST& \NaN& \NaN& \NaN& \NaN& \NaN& \NaN\\
    20& 2022-01-01 10:07 EST& \NaN& \NaN& \NaN& \NaN& \NaN& \NaN\\
    \hline
  \end{tabular}
\end{table}

The data from the ``PRE-OPERATIVE~MANAGMENT'' portion of the DCD~Flowsheet form
require a second table. Thus, one relatively short form may leads to 2
or more tables in the target database.  This table has a large number
of mostly categorical variables and is best presented in a format which
lists variables vertically rather than horizontally, such as JSON
shown in Table~\ref{tbl:preop-section-json}. Another objective
of generating JSON is to confirm that our data can be stored in JSON-oriented
databases, such as MongoDB.

It should be noted that the data extracted from the PDF attachements is captured in
native MATLAB datastructures, which is the directly produced
intermediate format resulting from MATLAB programming. Subsequently,
APIs have been constructed, so that we can export the data in various
formats. The format of the exported data is compatible with
industry-standard RDBMS systems and, at the time of this writing, the
relational database consists of 18 tables, quickly growing
as new types of medical forms are incorporated. Currently implemented
forms are identified by their titles:
\begin{enumerate}
\item ``DCD FLOWSHEET''
\item ``PRE-OPERATIVE~MANAGMENT'' (separate table holding a portion of
  the ``DCD Flowsheet'')
\item ``FLOWSHEET''
\item ``KIDNEY PERFUSION FLOW SHEET''
\item ``LIVER DATA''
\item ``REFERRAL WORKSHEET'' 
\end{enumerate}
The form with title ``FLOWSHEET'' is actually a collection of subforms, which
we mapped to separate database tables:
\begin{enumerate}
\item ``VITAL SIGNS''
\item ``VENT SETTINGS''
\item ``INTAKE''
\item ``Medications Dosage''
\item ``OUTPUT''
\item ``Comments''
\end{enumerate}
The remaining tables consist of book-keeping data, such as patient information, 
document version information, and timestamps when document was generated.

This data is then \emph{exported} using MATLAB built-in interfaces to the following data formats
\begin{enumerate}
\item SQL database (we're using SQLite via a JDBC driver).
\item MongoDB using MATLAB wrapper around 'mongoc', which is a C-based
  API for MongoDB, and JSON generated by MATLAB command 'jsonencode'.
\item Plain SQL file obtained by exporting the SQLite database to SQL (equivalent
  to a complete database backup).
\item CSV, suitable for a quick export of individual data tables.
\end{enumerate}

\section{Highlights of processing}
\href{https://soteria.arizona.edu/}{Soteria} is a HIPAA-compliant HPC
system, which holds the OPTN dataset (HIPAA-compliance is
mandated). The system features 96 CPU and GPU (currently not used in
our project), running a flavor of Linux. The software is build on top
of MATLAB and its add-on packages called Toolboxes. Full version of
MATLAB with toolboxes is a rich programming environment.  When
installed, it occpuies 20-30 GB of disk space.  We call Java
libraries, such as \emph{PDFBox} and in-house built Java packages,
from MATLAB, thanks to a very functional MATLAB-Java interface.

As an example, as 36-page PDF attachment is processed in $4.6$ seconds
on a single CPU. This size is representative of the avarage size of an
attachment, but they widely vary in size. Thus, the 2022 dataset may
take $4\times 10^5\times 4.6=1,840,000$ seconds, or $511$ hours. Given
that files can be processed in parallel (``trivial parallelism'') the
real time of processing can be reduced to as little as $511/96=5.324$
hours. In practice, the time is approximately $1/3$ of that because
many attachments can be skipped (e.g., image-based PDFs).

It is clear that software performance is not an issue, as the entire
national organ transplant dataset can be processed in $15\times 5=75$
hours, i.e. roughly 3 days. Hence, our system architecture is suited
to fully addressing the problem at hand.

\section{Image-based PDF}
The methodology for image-based PDF is different, as it required OCR.
OCR, by its nature, does not recover text with absolute accuracy, but
our preliminary results indicate that, combined with NLP techniques
(e.g., a custom spell-checker) and custom training on the fonts that
are actually used in the OPTN forms, we demonstrated near-100\%
accuracy on selected forms, e.g, ``RENAL DATA'' which contains
detailed kidney anatomy information. In future papers we will cover
image-based PDF in more detail.

\section{Large Language Models}
Large Language Models (LLM) are gaining prominence due to the
launch of ChatGPT. The data extracted from PDF contains blocks
of free-form text containing valuable information. Incorporation
of LLMs into our software is ongoing, and will be present the results
in future papers.

\section{Conclusions}

In U.S. Donor data has been stored in OPTN's DonorNet as each donor's
profile as individual PDF attachments since 2008. With 57 OPOs using
over 25 different forms to capture kidney anatomy and pathology
reporting, Over 1 million PDF attachments for 135,000 deceased donors
covering 2008 to 2021 are identified.

The article discusses a methodology for extracting and analyzing data
from the PDF attachments, specifically focusing on medical forms like
DCD flowsheets. The process involves converting the PDF documents into
analyzable data formats, which are then subjected to image processing
and regular grammar-based parsing.

\vskip2em
\noindent\textbf{Acknowledgement:} The data reported here been supplied
by UNOS as the contractor for the Organ Procurement and
Transplantation Network (OPTN). The interpretation and reporting of
these data are the responsibility of the author(s) and in no way
should be seen as an official policy of or interpretation by the OPTN
or the U.S. Government.

This study used data from the Organ Procurement and Transplantation
Network (OPTN). The OPTN data system includes data on all donor,
wait-listed candidates, and transplant recipients in the US, submitted
by the members of the Organ Procurement and Transplantation Network
(OPTN). The Health Resources and Services Administration (HRSA),
U.S. Department of Health and Human Services provides oversight to the
activities of the OPTN contractor.

\section*{References}
%\label{references}

% Rename the bibliography section.
% \bibliographystyle{unsrt}
\renewcommand{\refname}{Bibliography \& References Cited}
% \emergencystretch=1em
\printbibliography

@Misc{article1-2023,
  key =		 {art1-2023},
  author =	 {Marek Rychlik, Bekir Tanriover, Yan Han},
  title =	 {Extracting Semantics with text from PDFs},
  howpublished = {In preparation},
  year =	 2023,
}

@article{Keijbeck_Veenstra_Pol_Konijn_Jansen_vanGoor_Hoitsma_Peutz-Kootstra_Moers_2020,
  address =	 {United States},
  title =	 {The Association Between Macroscopic Arteriosclerosis
                  of the Renal Artery, Microscopic Arteriosclerosis,
                  Organ Discard, and Kidney Transplant Outcome},
  volume =	 {104},
  ISSN =	 {0041-1337},
  DOI =		 {10.1097/TP.0000000000003189},
  number =	 {12},
  journal =	 {Transplantation},
  publisher =	 {Copyright Wolters Kluwer Health, Inc. All rights
                  reserved},
  author =	 {Keijbeck, Anke and Veenstra, Rob and Pol, Robert A
                  and Konijn, Cynthia and Jansen, Nichon and van Goor,
                  Harry and Hoitsma, Anies J and Peutz-Kootstra,
                  Carine J and Moers, Cyril},
  year =	 {2020},
  pages =	 {2567–2574},
  language =	 {eng}
}

@article{Stewart_Foutz_Kamal_Weiss_McGehee_Cooper_Gupta_2022,
  title =	 {The Independent Effects of Procurement Biopsy
                  Findings on 10-Year Outcomes of Extended Criteria
                  Donor Kidney Transplants},
  volume =	 {7},
  ISSN =	 {2468-0249},
  DOI =		 {10.1016/j.ekir.2022.05.027},
  number =	 {8},
  journal =	 {Kidney international reports},
  publisher =	 {Elsevier Inc},
  author =	 {Stewart, Darren E. and Foutz, Julia and Kamal, Layla
                  and Weiss, Samantha and McGehee, Harrison S. and
                  Cooper, Matthew and Gupta, Gaurav},
  year =	 {2022},
  pages =	 {1850–1865},
  language =	 {eng}
}

@article{Gerber_Arrington_Taranto_Baker_Sung_2010,
  address =	 {Malden, USA},
  title =	 {DonorNet and the Potential Effects on Organ
                  Utilization},
  volume =	 {10},
  ISSN =	 {1600-6135},
  DOI =		 {10.1111/j.1600-6143.2010.03036.x},
  note =	 {Note on sources: The articles in this report are
                  based on the reference tables in the 2009 OPTN/SRTR
                  Annual Report. Table numbers are noted in brackets
                  and may be found online at},
  number =	 {4p2},
  journal =	 {American journal of transplantation},
  publisher =	 {Blackwell Publishing},
  author =	 {Gerber, D. A and Arrington, C. J and Taranto, S. E
                  and Baker, T and Sung, R. S},
  year =	 {2010},
  pages =	 {1081–1089},
  language =	 {eng}
}

@misc{The_UNOS_STAR_Deceased_Donor_SAF_file,
  title =	 {The UNOS STAR Deceased Donor SAF file},
  note =	 {accessed on 9/21/2022}
}

@misc{Public_Law_98_507,
  title =	 {Public Law 98-507-Oct.19, 1984. The National Organ Transplant Act},
  url =		 {https://www.congress.gov/98/statute/STATUTE-98/STATUTE-98-Pg2339.pdf},
  note =	 {Accessed on 9/24/2022}
}

@article{Massie_Zeger_Montgomery_Segev_2009,
  address =	 {Malden, USA},
  title =	 {The Effects of DonorNet 2007 on Kidney Distribution Equity and Efficiency},
  volume =	 {9},
  ISSN =	 {1600-6135},
  DOI =		 {10.1111/j.1600-6143.2009.02670.x},
  number =	 {7},
  journal =	 {American journal of transplantation},
  publisher =	 {Blackwell Publishing},
  author =	 {Massie, A. B and Zeger, S. L and Montgomery, R. A and Segev, D. L},
  year =	 {2009},
  pages =	 {1550–1557},
  language =	 {eng}
}

@article{Tierie_Roodnat_Dor_2019,
  title =	 {Systematic Surgical Assessment of Deceased-Donor Kidneys as a Predictor of Short-Term Transplant
                  Outcomes},
  volume =	 {60},
  ISSN =	 {0014-312X},
  number =	 {3-4},
  journal =	 {European surgical research},
  author =	 {Tierie, EL and Roodnat, J.I and Dor, F},
  year =	 {2019},
  pages =	 {97–105},
  language =	 {eng}
}

@article{Woestenburg_Sennesael_Bosmans_Verbeelen_2008,
  address =	 {United States},
  title =	 {Vasculopathy in the kidney allograft at time of transplantation: impact on later function of the
                  graft},
  volume =	 {85},
  ISSN =	 {0041-1337},
  DOI =		 {10.1097/TP.0b013e318169c311},
  number =	 {7 Suppl},
  journal =	 {Transplantation},
  author =	 {Woestenburg, Annemie and Sennesael, Jacques and Bosmans, Jean-Louis and Verbeelen, Dierik},
  year =	 {2008},
  pages =	 {S10–S18},
  language =	 {eng}
}

@article{Husain_King_Robbins-Juarez_Adler_McCune_Mohan_2021,
  address =	 {United States},
  title =	 {Number of Donor Renal Arteries and Early Outcomes after Deceased Donor Kidney Transplantation},
  volume =	 {2},
  ISSN =	 {2641-7650},
  DOI =		 {10.34067/KID.0005152021},
  number =	 {11},
  journal =	 {Kidney360},
  author =	 {Husain, S Ali and King, Kristen L and Robbins-Juarez, Shelief and Adler, Joel T and McCune, Kasi R and
                  Mohan, Sumit},
  year =	 {2021},
  pages =	 {1819–1826},
  language =	 {eng}
}

@article{Dare_Pettigrew_Saeb-Parsy_2014,
  address =	 {United States},
  title =	 {Preoperative Assessment of the Deceased-Donor Kidney: From Macroscopic Appearance to Molecular
                  Biomarkers},
  volume =	 {97},
  ISSN =	 {0041-1337},
  DOI =		 {10.1097/01.TP.0000441361.34103.53},
  number =	 {8},
  journal =	 {Transplantation},
  publisher =	 {by Lippincott Williams & Wilkins},
  author =	 {Dare, Anna J and Pettigrew, Gavin J and Saeb-Parsy, Kourosh},
  year =	 {2014},
  pages =	 {797–807},
  language =	 {eng}
}

@article{Gill_Rose_Lesage_Joffres_Gill_OConnor_2017,
  address =	 {United States},
  title =	 {Use and Outcomes of Kidneys from Donation after Circulatory Death Donors in the United States},
  volume =	 {28},
  ISSN =	 {1046-6673},
  DOI =		 {10.1681/ASN.2017030238},
  number =	 {12},
  journal =	 {Journal of the American Society of Nephrology},
  publisher =	 {American Society of Nephrology},
  author =	 {Gill, John and Rose, Caren and Lesage, Julie and Joffres, Yayuk and Gill, Jagbir and O'Connor, Kevin},
  year =	 {2017},
  pages =	 {3647–3657},
  language =	 {eng}
}

@article{Brennan_Sandoval_Husain_King_Dube_Tsapepas_Mohan_Ratner_2020,
  address =	 {Denmark},
  title =	 {Impact of warm ischemia time on outcomes for kidneys donated after cardiac death Post‐KAS},
  volume =	 {34},
  ISSN =	 {0902-0063},
  DOI =		 {10.1111/ctr.14040},
  number =	 {9},
  journal =	 {Clinical transplantation},
  publisher =	 {Wiley Subscription Services, Inc},
  author =	 {Brennan, Corey and Sandoval, Pedro Rodrigo and Husain, Syed Ali and King, Kristen L. and Dube,
                  Geoffrey K. and Tsapepas, Demetra and Mohan, Sumit and Ratner, Lloyd E.},
  year =	 {2020},
  pages =	 {e14040–n/a},
  language =	 {eng}
}

@article{Wang_Wetmore_Crary_Kasiske_2015,
  address =	 {United States},
  title =	 {The Donor Kidney Biopsy and Its Implications in Predicting Graft Outcomes: A Systematic Review},
  volume =	 {15},
  ISSN =	 {1600-6135},
  DOI =		 {10.1111/ajt.13213},
  number =	 {7},
  journal =	 {American journal of transplantation},
  publisher =	 {Wiley Subscription Services, Inc},
  author =	 {Wang, C. J. and Wetmore, J. B. and Crary, G. S. and Kasiske, B. L.},
  year =	 {2015},
  pages =	 {1903–1914},
  language =	 {eng}
}

@article{Lentine_Kasiske_Axelrod_2021,
  title =	 {Procurement Biopsies in Kidney Transplantation: More Information May Not Lead to Better Decisions},
  volume =	 {32},
  ISSN =	 {1046-6673},
  DOI =		 {10.1681/ASN.2021030403},
  number =	 {8},
  journal =	 {Journal of the American Society of Nephrology},
  author =	 {Lentine, Krista L. and Kasiske, Bertram and Axelrod, David A.},
  year =	 {2021},
  pages =	 {1835–1837},
  language =	 {eng}
}

@article{Husain_Shah_AlvaradoVerduzco_King_Brennan_Batal_Coley_Hall_Stokes_Dube_etal._2020,
  address =	 {United States},
  title =	 {Impact of Deceased Donor Kidney Procurement Biopsy Technique on Histologic Accuracy},
  volume =	 {5},
  ISSN =	 {2468-0249},
  DOI =		 {10.1016/j.ekir.2020.08.004},
  number =	 {11},
  journal =	 {Kidney international reports},
  publisher =	 {Elsevier Inc},
  author =	 {Husain, S. Ali and Shah, Vaqar and Alvarado Verduzco, Hector and King, Kristen L. and Brennan, Corey
                  and Batal, Ibrahim and Coley, Shana M. and Hall, Isaac E. and Stokes, M. Barry and Dube, Geoffrey
                  K. and Crew, R. John and Perotte, Adler and Natarajan, Karthik and Carpenter, Dustin and Sandoval,
                  P. Rodrigo and Santoriello, Dominick and D’Agati, Vivette and Cohen, David J. and Ratner, Lloyd and
                  Markowitz, Glen and Mohan, Sumit},
  year =	 {2020},
  pages =	 {1906–1913},
  language =	 {eng}
}

@article{Carpenter_Husain_Brennan_Batal_Hall_Santoriello_Rosen_Crew_Campenot_Dube_etal._2018,
  address =	 {United States},
  title =	 {Procurement Biopsies in the Evaluation of Deceased Donor Kidneys},
  volume =	 {13},
  ISSN =	 {1555-9041},
  DOI =		 {10.2215/CJN.04150418},
  note =	 {D.C. and S.A.H. contributed equally to this work.},
  number =	 {12},
  journal =	 {Clinical journal of the American Society of Nephrology},
  publisher =	 {American Society of Nephrology},
  author =	 {Carpenter, Dustin and Husain, S Ali and Brennan, Corey and Batal, Ibrahim and Hall, Isaac E and
                  Santoriello, Dominick and Rosen, Raphael and Crew, R John and Campenot, Eric and Dube, Geoffrey K and
                  Radhakrishnan, Jai and Stokes, M Barry and Sandoval, P Rodrigo and D'Agati, Vivette and Cohen, David J
                  and Ratner, Lloyd E and Markowitz, Glen and Mohan, Sumit},
  year =	 {2018},
  pages =	 {1876–1885},
  language =	 {eng}
}

@article{Lentine_Naik_Schnitzler_Randall_Wellen_Kasiske_Marklin_Brockmeier_Cooper_Xiao_etal._2019,
  address =	 {United States},
  title =	 {Variation in use of procurement biopsies and its implications for discard of deceased donor kidneys
                  recovered for transplantation},
  volume =	 {19},
  ISSN =	 {1600-6135},
  DOI =		 {10.1111/ajt.15325},
  note =	 {This study was approved by the Saint Louis University Institutional Review Board. Individual
                  participant deidentified data will not be shared by the authors due to restrictions of Data Use
                  Agreements. SRTR registry data can obtained from the SRTR.},
  number =	 {8},
  journal =	 {American journal of transplantation},
  publisher =	 {Wiley Subscription Services, Inc},
  author =	 {Lentine, Krista L. and Naik, Abhijit S. and Schnitzler, Mark A. and Randall, Henry and Wellen, Jason
                  R. and Kasiske, Bertram L. and Marklin, Gary and Brockmeier, Diane and Cooper, Matthew and Xiao,
                  Huiling and Zhang, Zidong and Gaston, Robert S. and Rothweiler, Richard and Axelrod, David A.},
  year =	 {2019},
  pages =	 {2241–2251},
  language =	 {eng}
}

@article{Tingle_Figueiredo_Moir_Goodfellow_Talbot_Wilson_Tingle_2019,
  address =	 {Chichester, UK},
  title =	 {Machine perfusion preservation versus static cold storage for deceased donor kidney transplantation},
  volume =	 {2019},
  ISSN =	 {1465-1858},
  DOI =		 {10.1002/14651858.CD011671.pub2},
  note =	 {Editorial Group: Cochrane Kidney and Transplant Group.},
  number =	 {3},
  journal =	 {Cochrane database of systematic reviews},
  publisher =	 {John Wiley & Sons, Ltd},
  author =	 {Tingle, Samuel J and Figueiredo, Rodrigo S and Moir, John AG and Goodfellow, Michael and Talbot, David
                  and Wilson, Colin H and Tingle, Samuel J},
  year =	 {2019},
  pages =	 {CD011671–CD011671},
  language =	 {eng}
}

@misc{pdflayouttextstripper,
  author =	 {Jonathan Link},
  title =	 {PDFLayoutTextStripper},
  url =		 {https://jonathanlink.ch/PDFLayoutTextStripper.html},
  year =	 {2022},
}

@misc{pdfbox,
  author =	 {The Apache Software Foundation},
  title =	 {Apache PDFBox\textsuperscript{\textregistered} - A Java PDF Library},
  url =		 {https://pdfbox.apache.org/},
  year =	 2022
}

\end{document}